\documentclass[conference]{IEEEtran}
\usepackage[utf8]{inputenc}
\usepackage[T1]{fontenc}
\usepackage{amssymb}

\ifCLASSINFOpdf
\else
\fi
\usepackage{amsmath}
\usepackage{algorithmic}
\usepackage{algorithm}

\usepackage{mdwmath}
\usepackage{mdwtab}

\ifCLASSOPTIONcompsoc
    \usepackage[caption=false,font=normalsize,labelfont=sf,textfont=sf]{subfig}
\else
    \usepackage[caption=false,font=footnotesize]{subfig}
\fi

\usepackage{xcolor}
\usepackage{scalerel,verbatimbox}

\def\mycircle{\scalerel*{\addvbuffer[-.15pt -.6pt]{$\circ$}}{\blacksquare}}

\newcommand{\bluecircle}{\textcolor{blue}{\mycircle}}
\newcommand{\redsquare}{\textcolor{red}{\blacksquare}}

\begin{document}
%
\title{K-Nearest Oracles Borderline\\Dynamic Classifier Ensemble Selection}


\author{\IEEEauthorblockN{
    Dayvid V. R. Oliveira\IEEEauthorrefmark{1},
    George D. C. Cavalcanti\\
    and
    Thyago N. Porpino
    }
    \IEEEauthorblockA{
        Centro de Inform\'atica - Universidade Federal de Pernambuco\\
        Email: \{dvro, gdcc, tnp\}@cin.ufpe.br
    }
    \and
    \IEEEauthorblockN{
        Rafael M. O. Cruz
        and
        Robert Sabourin
    }
    \IEEEauthorblockA{
        École de Technologie Sup\'erieure - Universit\'e du Qu\'ebec\\
        Email: \{cruz, robert.sabourin\}@livia.etsmtl.ca
    }
}


%


\maketitle


\begin{abstract}
    Dynamic Ensemble Selection (DES) techniques aim to select locally competent classifiers for the classification of each new test sample. Most DES techniques estimate the competence of classifiers using a given criterion over the region of competence of the test sample (its the nearest neighbors in the validation set). The K-Nearest Oracles Eliminate (KNORA-E) DES selects all classifiers that correctly classify all samples in the region of competence of the test sample, if such classifier exists, otherwise, it removes from the region of competence the sample that is furthest from the test sample, and the process repeats. When the region of competence has samples of different classes, KNORA-E can reduce the region of competence in such a way that only samples of a single class remain in the region of competence, leading to the selection of locally incompetent classifiers that classify all samples in the region of competence as being from the same class. In this paper, we propose two DES techniques: K-Nearest Oracles Borderline (KNORA-B) and K-Nearest Oracles Borderline Imbalanced (KNORA-BI). KNORA-B is a DES technique based on KNORA-E that reduces the region of competence but maintains at least one sample from each class that is in the original region of competence. KNORA-BI is a variation of KNORA-B for imbalance datasets that reduces the region of competence but maintains at least one minority class sample if there is any in the original region of competence. Experiments are conducted comparing the proposed techniques with 19 DES techniques from the literature using 40 datasets. The results show that the proposed techniques achieved interesting results, with KNORA-BI outperforming state-of-art techniques. 
\end{abstract}



%
\IEEEpeerreviewmaketitle

\section{Introduction}

Multiple Classifier Systems (MCS) \cite{mcs:2014} combine
classifiers in the hope that several classifiers outperform
any individual classifier in classification accuracy \cite{kuncheva:2004}.
MCS have been considered an interesting alternative for
increasing the classification accuracy in several studies
\cite{image_labeling:2005} \cite{handwritten:2013}
\cite{signature:2010} \cite{recommendation_systems:2010}
\cite{credit_card_fraud:2011} \cite{facerecognition:2015}
and machine learning competitions 
\cite{kaggle:2014} \cite{netflix:2009} \cite{netflix:2007}.

Dynamic Ensemble Selection (DES) \cite{des:2008} \cite{cruz2018dynamic} \cite{dcs:2014} techniques
select one or more classifiers for the classification of each new test
sample.
Relying on the assumption that different classifiers are
competent ("experts") in different local regions of the feature
space, most DES techniques estimate the level of competence of
a classifier for the classification of a test sample $x_{query}$,
using some criteria over the region of competence of $x_{query}$.
The region of competence of $x_{query}$ is the set of $K$ nearest
neighbors of $x_{query}$ in the validation set $\mathcal{D}_{SEL}$.

In \cite{des:2008}, Ko et al. proposed two DES techniques:
K-Nearest Oracles Eliminate (KNORA-E) and K-Nearest Oracles Union (KNORA-U).
KNORA-E selects all classifiers that correctly classify all samples in the region of competence
of a test sample. If no classifier is selected, KNORA-E removes from the region of competence
the sample that is furthest from the test sample until at least one classifier is selected.
KNORA-U selects all classifiers that correctly classify at least one sample in the region of competence,
the more samples a classifier correctly classifies, the more votes it has for the classification of
the test sample.

In \cite{dfp:2017}, Oliveira et al. showed that, when the
region of competence of a test sample is composed of samples
from different classes (indecision region),
DES techniques can select classifiers that classify all samples
in the region of competence as being from the same class.
The authors then proposed the Frienemy Indecision Region
Dynamic Ensemble Selection (FIRE-DES) framework.
This framework pre-selects classifiers with decision boundaries
crossing the region of competence of the test sample if the
test sample is located in an indecision region,
preventing DES techniques from selecting classifiers that
classify all samples in the region of competence to the
same class.

Considering KNORA-E, if no classifiers correctly classify all samples
in the region of competence of a test sample, KNORA-E can change a
region that was composed of samples from different classes
into a smaller region composed of samples of a single class.
FIRE-DES tackles this issue ensuring that, if the test sample
is located in an indecision region, KNORA-E will select only classifiers
with decision boundaries crossing the original region of competence.

Even though FIRE-DES tackles the indecision region problem of
KNORA-E, the way in which KNORA-E reduces the
region of competence can lead to the selection of incompetent
classifiers, even when using FIRE-DES, as the pre-selection
of classifiers is performed only once over the original region
of competence. In this paper, we propose two DES techniques:
K-Nearest Oracles Borderline (KNORA-B) and
K-Nearest Oracles Borderline Imbalanced (KNORA-BI).
KNORA-B is a DES technique based on KNORA-E that prevents
the underrepresentation of classes in the region of competence
when it is composed of samples from different classes.
KNORA-BI is a variation of KNORA-B for imbalanced datasets that
tackles the class imbalance problem \cite{cip:2013} by
preventing only the underrepresentation of the minority class
(class with few samples) in the region of competence
- but allowing the underrepresentation of the majority class
(class with many samples).

The remainder of this paper is organized as follows:
Section II presents the KNORA-E and KNORA-U.
Section III presents the problem statement.
Section IV presents the proposed techniques KNORA-B and KNORA-BI.
Section IV  presents the experiments.
Finally, Section V presents the conclusion.

\section{Background}

The Oracle concept is a hypothetical dynamic selection approach
that always selects the classifier that correctly classifies the
test sample, if such classifier exists \cite{kuncheva:2002}.

In \cite{des:2008}, Ko et al. introduced the concept of K-Nearest Oracles
and proposed two DES techniques: K-Nearest Oracles Union (KNORA-U) and
K-Nearest Oracles Eliminate (KNORA-E).
For the classification of a test sample, KNORA-E and KNORA-U find
the region of competence of the test sample and use the samples
in this region as \textit{oracles} to perform the selection
of classifiers (knowing the classifiers that correctly classify
each sample in the region of competence). The following subsection details the KNORA-Eliminate.

\subsection{KNORA-Eliminate}

Given a test sample $x_{query}$ to be classified,
KNORA-E finds the region of competence ($\Psi$) of $x_{query}$
by selecting the $K$ nearest neighbors of $x_{query}$ in the
validation set $\mathcal{D}_{SEL}$.
After that, KNORA-E selects all classifiers that correctly
classify all samples in $\Psi$.
If no classifiers correctly classify all samples in $\Psi$,
KNORA-E reduces $\Psi$ by removing the sample that is furthest
from $x_{query}$, until at least one classifier is selected.
If $\Psi$ gets empty, and no classifier was selected,
KNORA-E selects all classifiers with the same classification
accuracy as the single best classifier in the original $\Psi$.

\begin{figure}[h]
    \begin{center}
    \includegraphics[scale=0.32]{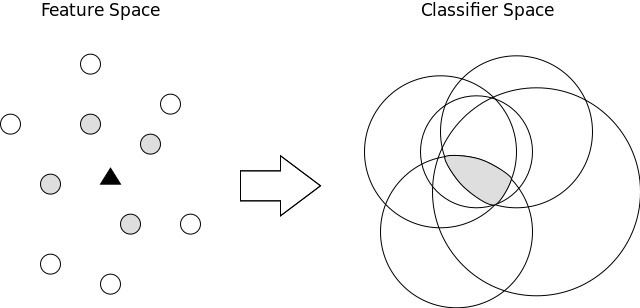}
    \caption{
        Selection of KNORA-E.
        On the left side, $\blacktriangle$ is the test sample, 
        and the darkened samples are the region of competence of the test sample.
        On the right side, the selected classifiers are darkened.
        (From \cite{des:2008}).
    }
    \label{fig:kne_ilustration}
    \end{center}
\end{figure}

Fig. \ref{fig:kne_ilustration} shows a test sample $\blacktriangle$,
its region of competence (darkened samples), and regions of expertise of
the classifiers (circles on the right side). KNORA-E selects
all classifiers that are experts for the classification of all
samples in the region of competence (intersection of correct classifiers).

\section{Problem Statement}
\label{sec:problem_statement}

When no classifier correctly classifies all samples in the region
of competence of a test sample, KNORA-E reduces the region of competence
by removing the sample that is the furthest from $x_{query}$, regardless of its class.
Because of that, KNORA-E can change a region of competence
that is composed of samples from different classes into a region of
competence composed of samples from a single class.

\begin{figure*}[!hbtp]
\begin{center}
    \subfloat[1st iteration of KNORA-E.]{
        \includegraphics[scale=0.325]{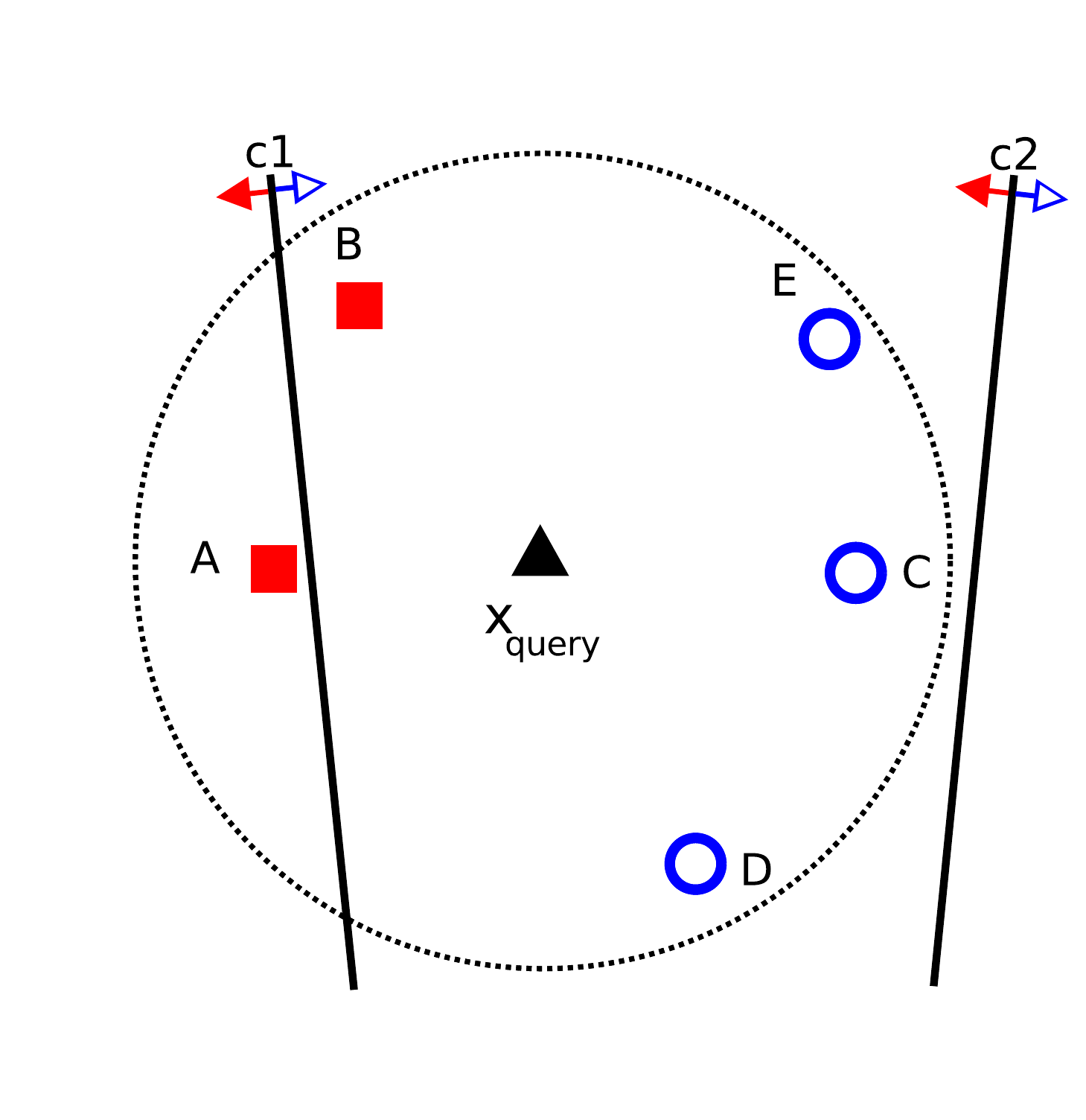}
        \label{subfig:kne1}
    }
    \hspace*{3cm}
    \subfloat[2nd iteration of KNORA-E.]{
        \includegraphics[scale=0.325]{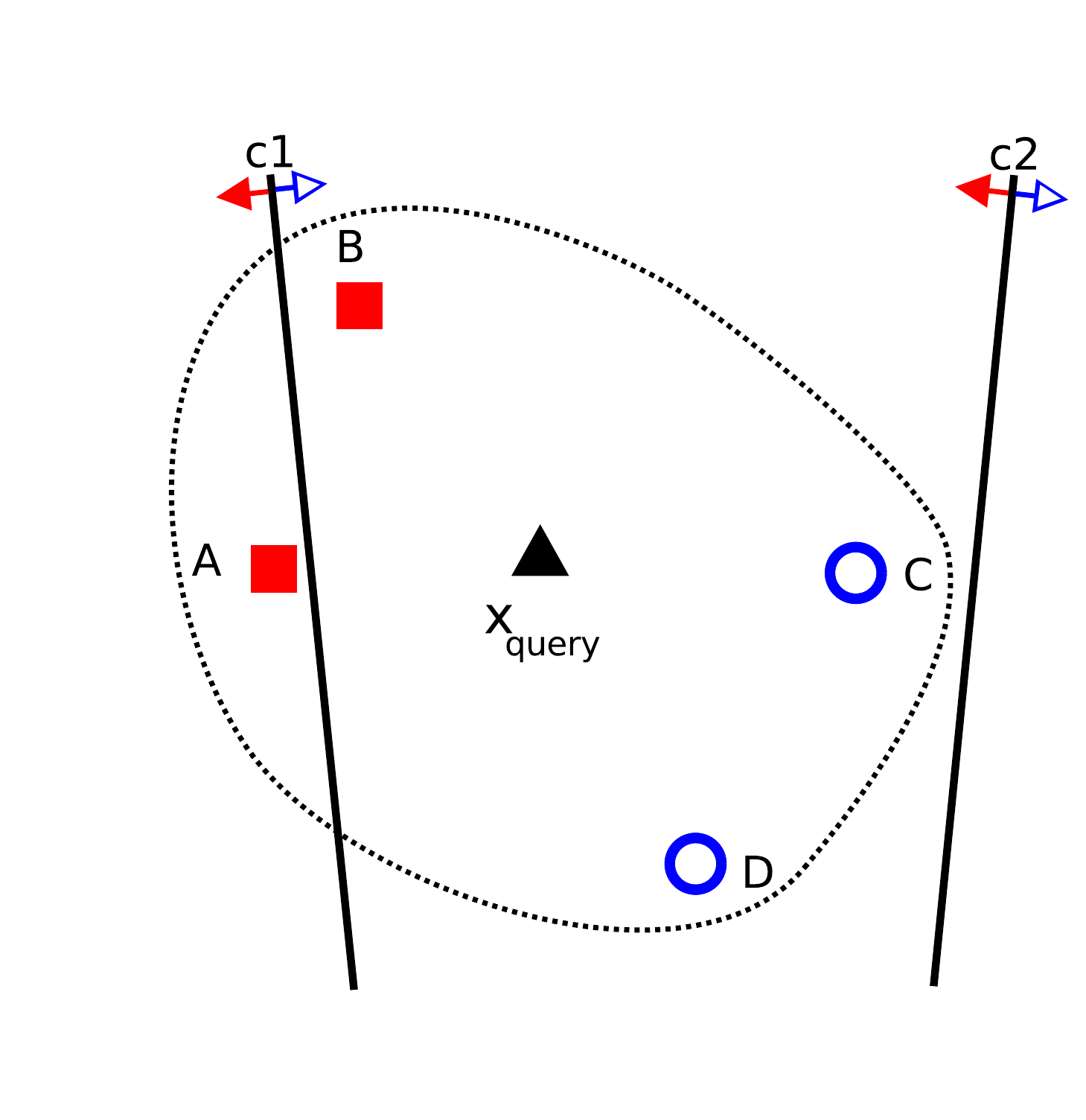}
        \label{subfig:kne2}
    }
    \hspace{10cm}
    \subfloat[3rd iteration of KNORA-E.]{
        \includegraphics[scale=0.325]{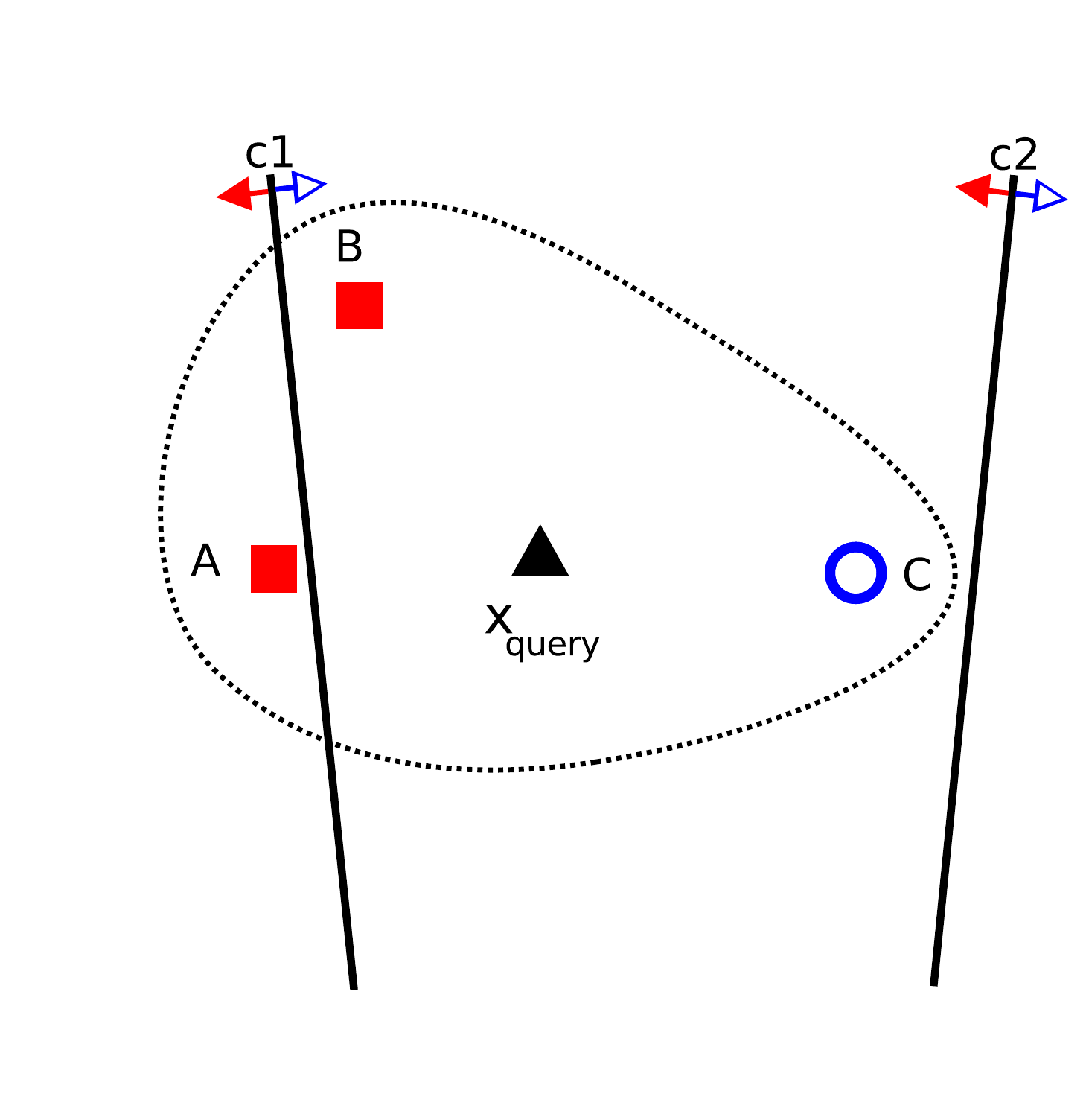}
        \label{subfig:kne3}
    }
    \hspace*{3cm}
    \subfloat[4th iteration of KNORA-E.]{
        \includegraphics[scale=0.325]{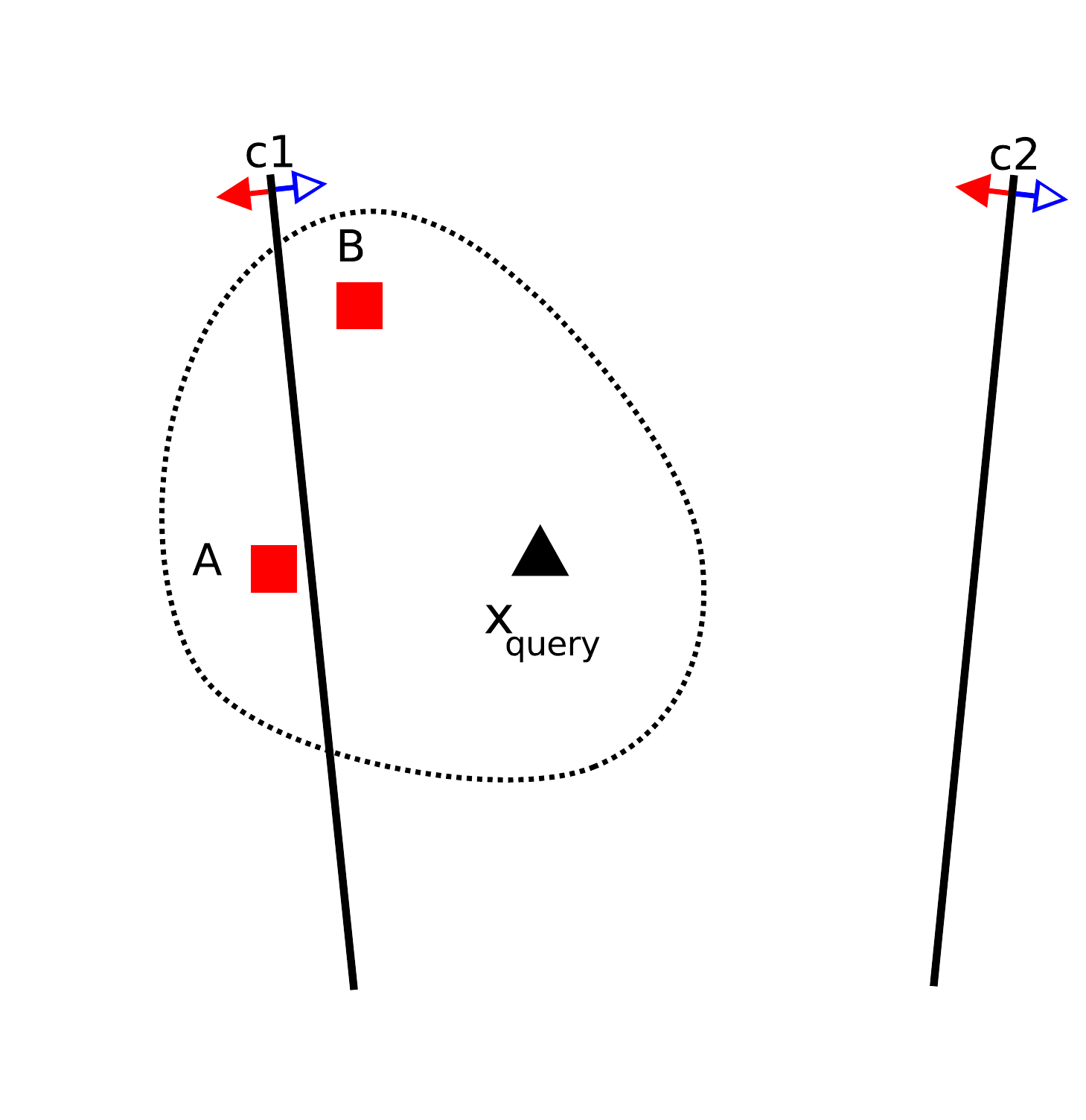}
        \label{subfig:kne4}
    }
    \caption{
        KNORA-E iterations when no classifier correctly classifies
        all samples in the region of competence 
        of the test sample.
        The $\blacktriangle$ is the test sample, 
        the markers {\large$\textcolor{blue}{\circ}$}
        and $\textcolor{red}{\blacksquare}$ are samples from different 
        classes in the region of competence of the test sample,
        and the class of the test sample is "{\large$\textcolor{blue}{\circ}$}".
        (Adapted from \cite{dfp:2017}).
    }
    \label{fig:kne}
\end{center}
\end{figure*}

Figure \ref{fig:kne} presents the iterations of KNORA-E ($K = 5$)
for the classification of a test sample ($x_{query}$) when no
classifier correctly classifies all samples in the region of
competence ($\Psi$) of $x_{query}$.
In this figure, $\blacktriangle$ is the test sample,
the dotted line delimits the region of competence,
$A$, $B$, $C$, $D$, and $E$ are the $K$ nearest neighbors of
$x_{query}$ in the validation set,
$c1$ and $c2$ are classifiers and
the continuous straight lines are their decision boundaries.
The markers $\bluecircle$ and $\redsquare$ represent samples from different classes,
and the test sample is from the class "$\bluecircle$".

In the scenario from Figure \ref{fig:kne},
in the first and second iterations (Figure \ref{subfig:kne1} and \ref{subfig:kne2})
no classifier correctly classifies all samples in $\Psi$, so KNORA-E removes
the samples $E$ and $D$ from $\Psi$, in the first and second iterations, respectively.

In the third iteration (Figure \ref{subfig:kne3}),
again, no classifier correctly classifies all samples in $\Psi$,
so KNORA-E removes the sample that is the furthest from $x_{query}$,
$C$, (the last remaining sample of class "$\bluecircle$" in $\Psi$), leaving only two
samples of the class "$\redsquare$" in $\Psi$.
In the fourth iteration (Figure \ref{subfig:kne4}), the classifier
$c2$ correctly classifies all samples in $\Psi$ ($A$ and $B$), so
KNORA-E selects $c2$, misclassifying the test sample.

This is an issue because when the region of competence of the test
sample has samples from different classes, KNORA-E may change the region
of competence in such a way that it is no longer a good representation
of the local region of the test sample.
This behavior is not ideal because classifiers that classify all
samples in the region of competence as "$\redsquare$" (such as $c2$)
are selected.
This is a problem especially when dealing with imbalanced dataset,
where a region of competence with samples from the minority class
(class with few samples) can be changed into a region with only
samples from the majority class (class with many samples), even
though the minority class is usually the class of interest.

\section{Proposed Techniques}

\subsection{KNORA-Borderline}

The $K$-Nearest Oracles-Borderline (KNORA-B) is a DES technique
based on KNORA-E that maintains the classes represented in the
original region of competence of the test sample when reducing
the region of competence.

Given a test sample $x_{query}$,
KNORA-B finds its region of competence ($\Psi$)
by selecting the $K$ nearest neighbors of $x_{query}$ in the
validation set $\mathcal{D}_{SEL}$.
Then, KNORA-B selects all classifiers that correctly
classify all $K$ samples in $\Psi$.
If no classifier correctly classifies all samples in $\Psi$,
KNORA-B reduces $\Psi$ by removing the sample that is the furthest
($\Psi_{b}$) from $x_{query}$, only if
all classes represented in $\Psi$ are still represented in $\Psi / \Psi_b$,
otherwise, KNORA-B evaluates the removal of the next furthest sample
($\Psi_{b-1}$).
The process repeats until at least one classifier
is selected. If KNORA-B reaches a state in which it is not possible
to remove any sample from $\Psi$ while maintaining the set of classes,
KNORA-B uses the KNORA-E rule over the original region of competence.

Algorithm \ref{alg:knb} presents the KNORA-B pseudocode.
$x_{query}$ is the test sample,
$C$ is the pool of classifiers,
$\mathcal{D}_{SEL}$ is the validation set,
and $K$ is the size of the region of competence.
First, KNORA-B initializes \textit{EoC} as an empty list
to insert the selected classifiers (Line 1), it then
selects the region of competence $\Psi$ of $x_{query}$
by applying the $K$-nearest neighbors on $\mathcal{D}_{SEL}$
(Line 2),
and stores this initial region of competence in $\Psi_{original}$
(Line 3).
Next, until at least one classifier is selected or until
the region of competence is empty (Line 4),
KNORA-B tries to select all classifiers that correctly classify
all samples in $\Psi$ (Line 5 - 9), if no classifier is selected
KNORA-B reduces the region of competence and the process repeats
(Lines 10 - 12).
If the region of competence can no longer be reduced, and no
classifier was selected, KNORA-B performs the \textit{fallback selection}
(Lines 14 - 16) - the fallback selection of KNORA-B is the KNORA-E procedure.
Finally, KNORA-B returns the selected ensemble of classifiers.

\begin{algorithm}[h]
\small
\caption{KNORA-B}
\begin{algorithmic}[1]
	\REQUIRE $\mathcal{C}$: pool of classifiers
    \REQUIRE $\mathcal{D}_{SEL}$: validation set
	\REQUIRE $x_{query}$: test sample
	\REQUIRE $K$: size of the region of competence
    \STATE \textit{EoC} $\gets $ ensemble of classifiers
    \STATE $\Psi \gets K$ nearest neighbors of $x_{query}$ in $\mathcal{D}_{SEL}$
    \STATE $\Psi_{original} \gets \Psi$
    \WHILE {Empty(\textit{EoC}) $\land$ $\neg$Empty($\Psi$)}
        \FORALL {$c_i$ in $\mathcal{C}$}
            \IF {$c_i$ correctly classifies all samples in $\Psi$}
                \STATE \textit{EoC} $\gets$ \textit{EoC} $\cup$ $c_i$
            \ENDIF
        \ENDFOR
        \IF {\textit{EoC} is empty}
            \STATE $\Psi \gets $\textit{reduced\_region\_of\_competence($x_{query}, \Psi$)}
        \ENDIF
    \ENDWHILE
    \IF {\textit{EoC} is empty}
    \STATE \textit{EoC} $\gets $ fallback\_selection($C, \Psi_{original}, x_{query}, K$)
    \ENDIF
	\RETURN \textit{EoC}
\end{algorithmic}
\label{alg:knb}
\end{algorithm}

KNORA-B differs from KNORA-E in the \textit{reduced region of competence}
procedure applied when no classifier correctly classifies all samples in
the region of competence ($\Psi$). Algorithm \ref{alg:knb_rroc} presents the region of competence reduction process.
Given a test sample $x_{query}$ and the region of competence $\Psi$,
KNORA-B gets the size $S$ of the region of competence (Line 1),
and assigns that to a variable $b$ (Line 2).
Now, until the region of competence is not reduced and $b$ is greater
than zero (Line 3), KNORA-B gets the b-th nearest sample
(starts with $b =$ \textit{SizeOf($\Psi$)}, that is, from the furthest to the nearest)
of $x_{query}$ ($\Psi_b$) in $\Psi$ (Line 4), and evaluates if
all classes represented in $\Psi$ are still represented if
$\Psi_b$ is removed from $\Psi$. If so, $\Psi_b$ is removed
from $\Psi$, otherwise, $b$ is decreased (Lines 5 - 9).
If no sample was removed from $\Psi$, this means that no reduction
was possible (there is only one sample from each class that
was represented in the original region of competence),
then $\Psi$ is assigned to an empty set (Lines 11 - 13)
so that KNORA-B can use the fallback.
Finally, the reduced region of competence is returned (Line 14).

\begin{algorithm}[h]
\small
\caption{KNORA-B - Reduced Region of Competence}
\begin{algorithmic}[1]
	\REQUIRE $x_{query}$: test sample
    \REQUIRE $\Psi$: region of competence of $x_{query}$
    \STATE $S \gets$ \textit{SizeOf($\Psi$)}
    \STATE $b \gets$ S
    \WHILE {\textit{SizeOf($\Psi$)} $ = S$ and $b > 0$}
        \STATE $\Psi_b \gets$ b-th nearest from $x_{query}$ in $\Psi$
        \IF {\textit{Set(classes($\Psi / \Psi_b$))} $=$ \textit{Set(classes($\Psi$))}}
            \STATE $\Psi \gets \Psi / \Psi_b$
        \ELSE
            \STATE $b \gets b - 1$
        \ENDIF
    \ENDWHILE
    \IF {$\textit{SizeOf($\Psi$)} = S$}
        \STATE $\Psi \gets \emptyset$
    \ENDIF
    \RETURN $\Psi$
\end{algorithmic}
\label{alg:knb_rroc}
\end{algorithm}

\begin{figure*}[!hbtb]
\centering
\subfloat[1st iteration of KNORA-B.]{
    \includegraphics[scale=0.325]{content/images/problem_statement/pdf/P1_A.pdf}
    \label{subfig:knb1}
}
\hspace*{3cm}
\subfloat[2nd iteration of KNORA-B.]{
    \includegraphics[scale=0.325]{content/images/problem_statement/pdf/P2_A.pdf}
    \label{subfig:knb2}
}
\hspace{10cm}
\subfloat[3rd iteration of KNORA-B.]{
    \includegraphics[scale=0.325]{content/images/problem_statement/pdf/P3_A.pdf}
    \label{subfig:knb3}
}
\hspace*{3cm}
\subfloat[4th iteration of KNORA-B.]{
    \includegraphics[scale=0.325]{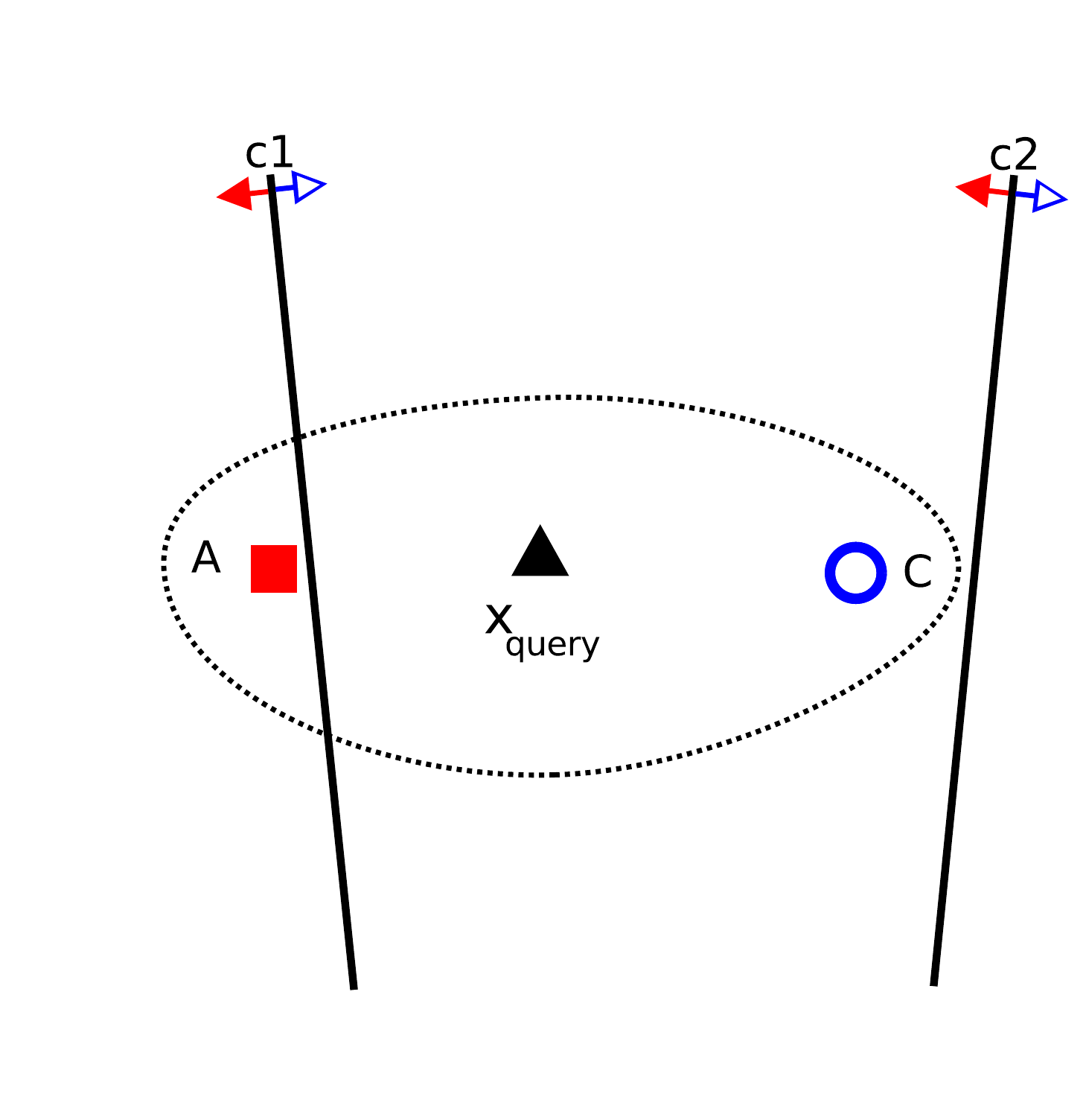}
    \label{subfig:knb4}
}
\caption{
        KNORA-B iterations when no classifier
        correctly classifies all samples in the region of competence 
        of the test sample.
        The $\blacktriangle$ is the test sample, 
        the markers {\large$\textcolor{blue}{\circ}$}
        and $\textcolor{red}{\blacksquare}$ are samples from different 
        classes in the region of competence of the test sample,
        and the class of the test sample is "{\large$\textcolor{blue}{\circ}$}".
        (Adapted from \cite{dfp:2017}).
}
\label{fig:knb}
\end{figure*}

Figure \ref{fig:knb} presents the iterations of KNORA-B ($K = 5$)
for the classification of a test sample ($x_{query}$) when no
classifiers correctly classify all samples in the region of
competence ($\Psi$) of $x_{query}$.
In this figure, $\blacktriangle$ is the test sample,
the dotted line delimits the region of competence,
$A$, $B$, $C$, $D$, and $E$ are the $K$ nearest neighbors of
$x_{query}$ in the validation set,
$c1$ and $c2$ are classifiers and
the continuous straight lines are their decision boundaries.
The markers $\bluecircle$ and $\redsquare$ are samples from different classes,
and the test sample is from the class "$\bluecircle$".

In the scenario from Figure \ref{fig:knb},
in the first and second iterations (Figure \ref{subfig:knb1} and \ref{subfig:knb2})
no classifier correctly classifies all samples in $\Psi$, so, KNORA-B removes
the sample that is the furthest from $x_{query}$, respectively, $E$ and $D$, from $\Psi$.
In the third iteration (Figure \ref{subfig:knb3}), again,
no classifier correctly classifies all samples in $\Psi$,
but instead of removing the furthest sample $C$, leaving
only samples from the class "$\redsquare$" in $\Psi$, KNORA-B
removes the
second furthest sample $B$, maintaining samples from both
classes "$\bluecircle$" and "$\redsquare$" in $\Psi$.
In the fourth iteration (Figure \ref{subfig:knb4}), the classifier
$c1$ correctly classifies all samples in $\Psi$ ($A$ and $C$), so,
KNORA-B selects $c1$ and correctly classifies the test sample
as being from the class "$\bluecircle$".

Comparing Figure \ref{fig:kne} with Figure \ref{fig:knb}, we can
see that KNORA-B is better in selecting classifiers for the 
classification of the test sample that has a region of competence
with samples of different classes.
While KNORA-E selects $c2$, a base classifier that classifies all samples
in $\Psi$ as being from the same class "$\redsquare$", and misclassifies
the test sample,
KNORA-B selects $c1$, a base classifier that correctly classifies samples from different classes
in $\Psi$, correctly classifying the test
sample.

\subsection{KNORA-Borderline-Imbalanced}

When dealing with specific problems such as imbalanced datasets,
KNORA-B can remove a minority class sample that is closer to the
test sample instead of a majority class sample that is the furthest.
This behavior is not desired in the context of imbalanced datasets,
in which the minority class (class with few samples) is the class
of interest \cite{cip:2013}.

KNORA-Borderline-Imbalanced (KNORA-BI) is a variation of KNORA-B
that has a different reduced region of competence reduction procedure
that allows the reduction of a region of competence that is composed
of samples from the majority and minority classes into a region of
competence composed only of samples from the minority class.
By doing so, KNORA-BI favors only the minority class when reducing
the region of competence.

KNORA-BI region of competence reduction pseudocode is presented
in Algorithm \ref{alg:knbi_rroc}. Given a test sample $x_{query}$
the region of competence $\Psi$, and the minority class class$_{min}$,
KNORA-BI gets the size $S$ of the region of competence (Line 1)
and assigns that to a variable $b$ (Line 2).
Now, until the region of competence is not reduced and $b$ is greater
than zero (Line 3), KNORA-BI gets the b-th nearest sample
(starts with $b =$ \textit{SizeOf($\Psi$)}, that is, from the furthest to the nearest)
of $x_{query}$ ($\Psi_b$) in $\Psi$ (Line 4),
if $\Psi_b$ is not from the minority class
or if all classes represented in $\Psi$ are still represented when
$\Psi_b$ is removed from $\Psi$, then, $\Psi_b$ is removed
from $\Psi$, otherwise, $b$ is decreased (Lines 5 - 9).
If no sample is removed from $\Psi$ and $b$ reaches zero,
$\Psi$ is assigned to an empty set (Lines 11 - 13). Thus, KNORA-BI can use the fallback.
Finally, the reduced region of competence is returned (Line 14).

\begin{algorithm}[h]
\small
\caption{KNORA-BI - Reduced Region of Competence}
\begin{algorithmic}[1]
	\REQUIRE $x_{query}$: test sample
    \REQUIRE $\Psi$: region of competence of $x_{query}$
    \REQUIRE class$_{min}$: minority class
    \STATE $S \gets$ \textit{SizeOf($\Psi$)}
    \STATE $b \gets$ S
    \WHILE {\textit{SizeOf($\Psi$)} $ = S$ and $b > 0$}
        \STATE $\Psi_b \gets$ b-th nearest from $x_{query}$ in $\Psi$
        \IF {Class($\Psi_b$) $\ne$ \textit{class}$_{min} \lor$ \textit{Set(classes($\Psi / \Psi_b$))} $=$ \textit{Set(classes($\Psi$))}}
            \STATE $\Psi \gets \Psi / \Psi_b$
        \ELSE
            \STATE $b \gets b - 1$
        \ENDIF
    \ENDWHILE
    \IF {$\textit{SizeOf($\Psi$)} = S$}
        \STATE $\Psi \gets \emptyset$
    \ENDIF
    \RETURN $\Psi$
\end{algorithmic}
\label{alg:knbi_rroc}
\end{algorithm}

Considering the examples from Figure \ref{fig:kne} and \ref{fig:knb},
KNORA-BI region of competence reduction procedure reduces the region
of competence in such a way that:
\begin{itemize}
    \item if the class "$\bluecircle$" is the majority class, KNORA-BI acts
        as exemplified in Figure \ref{fig:kne} - as it allows the removal of
        all majority class samples, leaving all minority class samples.
    \item if the class "$\bluecircle$" is the minority class, KNORA-BI acts
        as exemplified in Figure \ref{fig:knb} - as it does not allow the
        removal of all minority class samples.
\end{itemize}

\section{Experiments}
\label{sec:experiments}

We evaluated KNORA-B and KNORA-BI using 40 datasets as proposed
in \cite{dfp:2017}. The datasets were taken from the imbalanced datasets
module in the Knowledge Experiments based on Evolutionary Learning (KEEL)
repository \cite{keel}.
Table \ref{tab:datasets} presents the details about the datasets
used in our experiments: label, name, number of features,
number of samples, and imbalance ratio (IR).

\begin{table}[htbp]
\begin{center}
\caption{Summary of the 40 datasets used in the experiments: label, name, number of features, number of samples, and imbalance ratio (from \cite{dfp:2017}).}
\label{tab:datasets}
    \begin{tabular}{clrrr}
        \hline
        
Label & Name            & \#Feats. & \#Samples & IR \\
        \hline
        \hline
1     & glass1          & 9     & 214        & 1.82 \\
2     & ecoli0vs1       & 7     & 220        & 1.86 \\
3     & wisconsin       & 9     & 683        & 1.86 \\
4     & pima            & 8     & 768        & 1.87 \\ 
5     & iris0           & 4     & 150        & 2.00 \\
6     & glass0          & 9     & 214        & 2.06 \\
7     & yeast1          & 8     & 1484       & 2.46 \\ 
8     & vehicle2        & 18    & 846        & 2.88 \\ 
9     & vehicle1        & 18    & 846        & 2.9  \\
10    & vehicle3        & 18    & 846        & 2.99 \\ 
11    & glass0123vs456  & 9     & 214        & 3.2  \\ 
12    & vehicle0        & 18    & 846        & 3.25 \\ 
13    & ecoli1          & 7     & 336        & 3.36 \\ 
14    & new-thyroid1    & 5     & 215        & 5.14 \\ 
15    & new-thyroid2    & 5     & 215        & 5.14 \\
16    & ecoli2          & 7     & 336        & 5.46 \\
17    & segment0        & 19    & 2308       & 6.00 \\
18    & glass6          & 9     & 214        & 6.38 \\
19    & yeast3          & 8     & 1484       & 8.10 \\
20    & ecoli3          & 7     & 336        & 8.60 \\ 
21    & yeast-2vs4      & 8     & 514        & 9.08 \\
22    & yeast-05679vs4  & 8     & 528        & 9.35 \\
23    & vowel0          & 13    & 988        & 9.98 \\
24    & glass-016vs2    & 9     & 192        & 10.29 \\
25    & glass2          & 9     & 214        & 11.59 \\
26    & shuttle-c0vsc4  & 9     & 1829       & 13.87 \\
27    & yeast-1vs7      & 7     & 459        & 14.30 \\
28    & glass4          & 9     & 214        & 15.47 \\
29    & ecoli4          & 7     & 336        & 15.80 \\
30    & page-blocks-13vs4   & 10& 472        & 15.86 \\
31    & glass-0-1-6\_vs\_5  & 9 & 184        & 19.44 \\
32    & shuttle-c2-vs-c4    & 9 & 129        & 20.50 \\
33    & yeast-1458vs7   & 8     & 693        & 22.10 \\
34    & glass5          & 9     & 214        & 22.78 \\
35    & yeast-2vs8      & 8     & 482        & 23.10 \\
36    & yeast4          & 8     & 1484       & 28.10 \\
37    & yeast-1289vs7   & 8     & 947        & 30.57 \\
38    & yeast5          & 8     & 1484       & 32.73 \\
39    & ecoli-0137vs26  & 7     & 281        & 39.14 \\
40    & yeast6          & 8     & 1484       & 41.40 \\
        \hline
\end{tabular}
\end{center}
\end{table}

For each dataset, the data was partitioned using the
\textit{stratified 5-fold cross-validation}
(1 fold used for testing, and 4 folds for validation/training)
followed by a
\textit{stratified 4-fold cross-validation}
(the 4 folds into validation/training divided in 1 for
validation and 3 for training), resulting in 20
replications for each dataset using 20\% for testing,
20\% for validation, and 60\% for training.

The analysis is conducted using 8 DES techniques from the
literature, their respective FIRE-DES versions (using the F prefix), and
3 state-of-the-art DES techniques.
Table \ref{tab:dsexp} shows the dynamic selection techniques
used in our experiments, their categories and references.
Following the approach using in \cite{dfp:2017},
we use a pool of classifiers composed of 100 Perceptrons
generated using the Bootstrap AGGregatING (Bagging) technique \cite{bagging:1996},
and a region of competence size $K = 7$.

\begin{table}[h]
\begin{center}
\caption{
    Dynamic selection techniques considered in the experiments (From \cite{dfp:2017}).
}
\label{tab:dsexp}
\resizebox{0.450\textwidth}{!}{
\begin{tabular}{lll}
\hline
Technique & Category & Reference \\
\hline
\hline
\textbf{DES} & &  \\
Overall Local Accuracy (OLA)        & Accuracy      & Woods et al. \cite{dcs_la:1996} \\
Local Class Accuracy (LCA)          & Accuracy      & Woods et al. \cite{dcs_la:1996} \\
A Priori (APri)                     & Probabilistic &  Giacinto et al. \cite{dcs_la:1999} \\
A Posteriori (APos)                 & Probabilistic &  Giacinto et al. \cite{dcs_la:1999} \\
Multiple Classifier Behavior (MCB)  & Behavior      & Giacinto et al. \cite{mcb:2001} \\
Dynamic Selection KNN (DSKNN)       & Diversity     & Santana et al. \cite{dsknn:2006} \\
K-Nearests Oracles Union (KNORA-U)      & Oracle        & Ko et al. \cite{des:2008} \\
K-Nearests Oracles Eliminate (KNORA-E)  & Oracle        & Ko et al. \cite{des:2008} \\
\textbf{State-of-the-art} & & \\
Randomized Reference Classifier (RRC)   & Probabilistic &   Woloszynski et al. \cite{rrc:2011}      \\
META-DES                                & Meta-learning &   Cruz et al. \cite{metades:2015}         \\
META-DES.Oracle                         & Meta-learning &   Cruz et al. \cite{metadesoracle:2017}   \\
\hline
\end{tabular}}
\end{center}
\end{table}

Following the approach in \cite{dfp:2017}, 
we used the Area Under the ROC Curve (AUC) \cite{auc:1997}
for performance evaluation since
it is a suitable  metric for binary imbalanced
datasets \cite{cip:2013} \cite{dfp:2017}.
We also used the Wilcoxon Signed Rank Test \cite{wilcox:2016} \cite{wilcoxon:1945}
and the Sign Test \cite{signedtest:2006} to perform a pairwise comparison of the
proposed techniques with the techniques from the literature.

\subsection{Results}

Table \ref{tab:results} presents the overall results.
For each technique, the table shows:
the mean AUC and standard deviation,
the average ranking,
and the \textit{p-value} and result of the Wilcoxon Signed Rank Test
comparing KNORA-B and KNORA-BI with the DES technique
($+$/$=$/$-$ signs mean the proposed technique had statistically
better, equal, and worse classification performance
considering a confidence level $\alpha = 0.05$).

\begin{table}[h]
\begin{center}
\caption{
    Overall results.
}
\label{tab:results}
\resizebox{0.495\textwidth}{!}{
\begin{tabular}{lrrlrlr}
\hline
DES &              AUC &   RANK &      \multicolumn{2}{c}{KNORA-B (\textit{p-value})} &      \multicolumn{2}{c}{KNORA-BI (\textit{p-value})}  \\
\hline
\hline
           KNORA-BI &  \textbf{0.8136 (0.0743)} &         \textbf{6.80} &                0.9999 &          $-$ &  \multicolumn{2}{l}{$N/A$} \\
           META.O &  0.8067 (0.0649) &   8.16 &                0.9509 &          $-$ &                0.3454 &           $=$ \\
           META &  0.8100 (0.0635) &   8.20 &                0.9742 &          $-$ &                0.4362 &           $=$ \\
           FKNORA-U &  0.8081 (0.0765) &   8.38 &                0.9712 &          $-$ &                0.1009 &           $=$ \\
           FMCB &  0.8058 (0.0760) &   8.80 &                0.9235 &          $=$ &                0.0921 &           $=$ \\
           FKNORA-BI &  0.8083 (0.0758) &   9.01 &                0.9976 &          $-$ &                0.0003 &           $+$ \\
           FKNORA-E &  0.8055 (0.0768) &   9.59 &                0.9962 &          $-$ &                0.0002 &           $+$ \\
           FKNORA-B &  0.8042 (0.0772) &  10.34 &                0.9860 &          $-$ &                0.0001 &           $+$ \\
           KNORA-E &  0.8003 (0.0703) &  10.66 &                0.8298 &          $=$ &                0.0002 &           $+$ \\
           KNORA-B &  0.7989 (0.0721) &  11.28 &             \multicolumn{2}{l}{$N/A$} &  $5.60 \times e^{-6}$ &           $+$ \\
           FDSKNN &  0.8006 (0.0767) &  11.43 &                0.5250 &          $=$ &  $2.00 \times e^{-6}$ &           $+$ \\
           FLCA &  0.7946 (0.0777) &  11.84 &                0.5554 &          $=$ &  $8.30 \times e^{-6}$ &           $+$ \\
           RRC &  0.7934 (0.0658) &  12.26 &                0.1805 &          $=$ &                0.0057 &           $+$ \\
           FOLA &  0.8017 (0.0767) &  12.97 &                0.5026 &          $=$ &  $3.80 \times e^{-6}$ &           $+$ \\
           FAPRI &  0.7930 (0.0802) &  13.10 &                0.1940 &          $=$ &  $1.60 \times e^{-6}$ &           $+$ \\
           LCA &  0.7809 (0.0737) &  13.30 &                0.0332 &          $+$ &  $7.20 \times e^{-6}$ &           $+$ \\
           DSKNN &  0.7728 (0.0602) &  13.32 &                0.0107 &          $+$ &  $6.10 \times e^{-5}$ &           $+$ \\
           OLA &  0.7911 (0.0709) &  13.60 &                0.0623 &          $=$ &  $1.40 \times e^{-7}$ &           $+$ \\
           KNORA-U &  0.7560 (0.0548) &  15.39 &                0.0002 &          $+$ &  $6.80 \times e^{-6}$ &           $+$ \\
           FAPOS &  0.7681 (0.0809) &  15.86 &                0.0001 &          $+$ &  $1.40 \times e^{-7}$ &           $+$ \\
           APRI &  0.7537 (0.0628) &  16.38 &  $3.20 \times e^{-5}$ &          $+$ &  $9.80 \times e^{-7}$ &           $+$ \\
           APOS &  0.7380 (0.0658) &  18.01 &  $1.10 \times e^{-5}$ &          $+$ &  $4.70 \times e^{-7}$ &           $+$ \\
            MCB &  0.7443 (0.0566) &  17.32 &  $2.10 \times e^{-6}$ &          $+$ &  $6.40 \times e^{-7}$ &           $+$ \\
\hline
\end{tabular}}
\end{center}
\end{table}

Table \ref{tab:results} show that KNORA-BI achieved the highest
mean AUC ($0.8136$) and the best average ranking ($6.80$), outperforming all techniques considered in this work.
Moreover, it statistically outperformed 18 out of 22 techniques according to the Wilcoxon Signed Rank test.
Table \ref{tab:results} also shows that KNORA-B was not as good as KNORA-BI,
achieving the 12th best AUC ($0.7989$), and being statistically equivalent to KNORA-E ($0.8003$). However,
it was only statistically outperformed by 7 out of 22 techniques, where 3 of those are
variations of the techniques proposed in this paper (KNORA-BI, FKNORA-B, and FKNORA-BI).

In addition, FKNORA-B ($0.8042$) outperformed KNORA-B ($0.7989$) with statistical confidence,
meaning FIRE-DES caused a significant increase in classification performance in KNORA-B.
This increase in AUC is explained in the scenario that no classifier was selected
and the region of competence has only one sample from each class remaining.
In this scenario, KNORA-B applies KNORA-E on the entire pool of classifiers
while FKNORA-B applies KNORA-E on the set of pre-selected classifiers
(avoiding the selection of incompetent classifiers that classify all samples
in the region of competence as being from the same class).

On the other hand, KNORA-BI ($0.8136$) outperformed FKNORA-BI ($0.8083$) with statistical confidence,
meaning FIRE-DES caused a significant decrease in classification performance in KNORA-BI.
KNORA-BI allows the region of competence to be reduced until there is
only minority class samples, allowing to the selection of classifiers that classify all samples in
the region of competence as minority class samples, while FKNORA-BI does not allow the selection of
such classifiers. Because the minority class is so rare, preventing a DES technique
from selecting a classifier that classifies all samples in the region of competence as minority
class sample is not advantageous, specially because correctly classifying 1 minority class sample
has a higher impact than misclassifying 1 majority class sample,
which explains why KNORA-BI was better without FIRE-DES.

\begin{figure*}[!hbtp]
\begin{center}
    \subfloat[Comparison of KNORA-B and other techniques.]{
        \includegraphics[width=2.6in, height=4.8in]{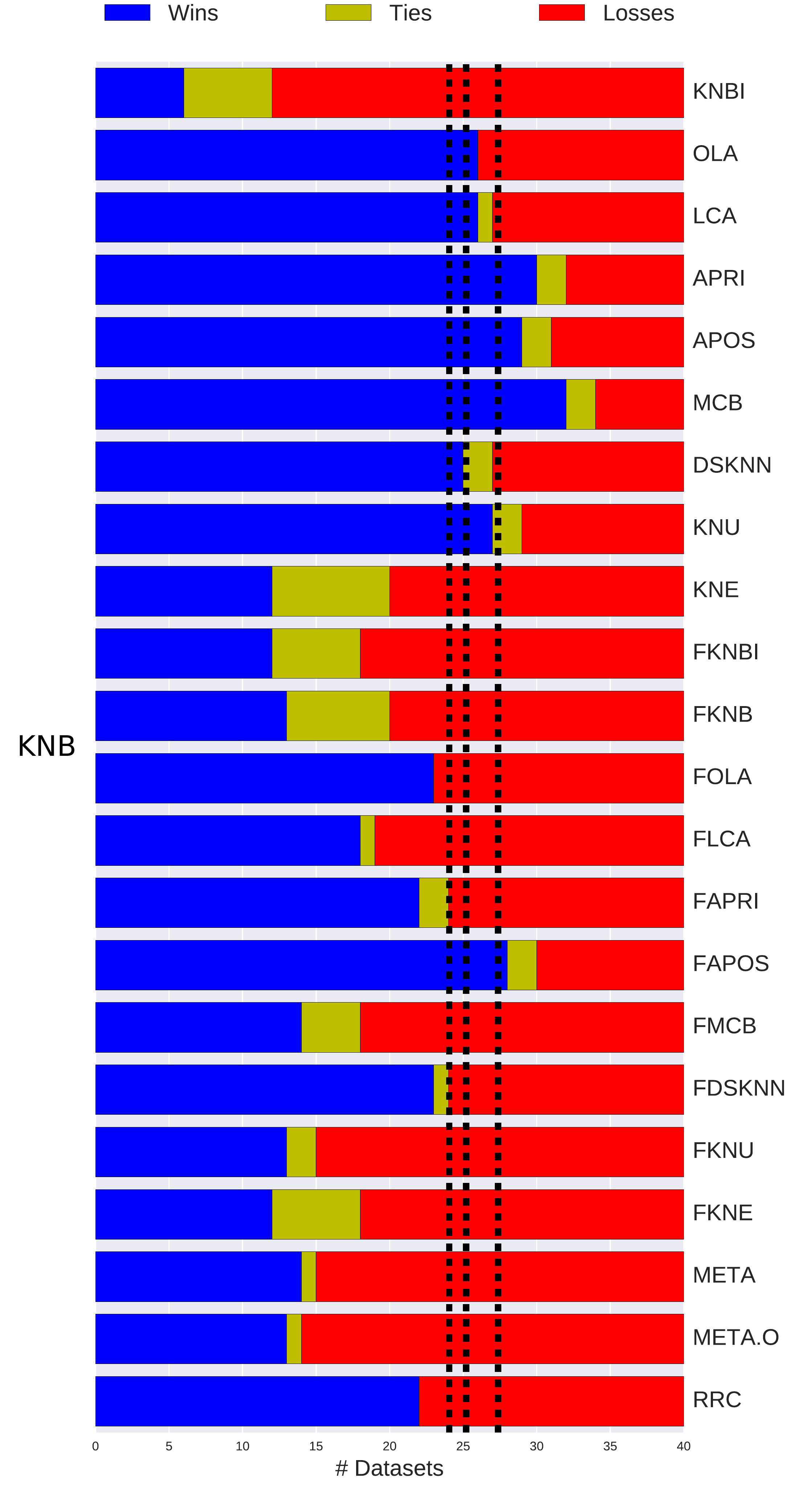}
        \label{fig:wtl_knb_all}
    }
    \hspace*{3cm}
    \subfloat[Comparison of KNORA-BI and other techniques.]{
        \includegraphics[width=2.6in, height=4.8in]{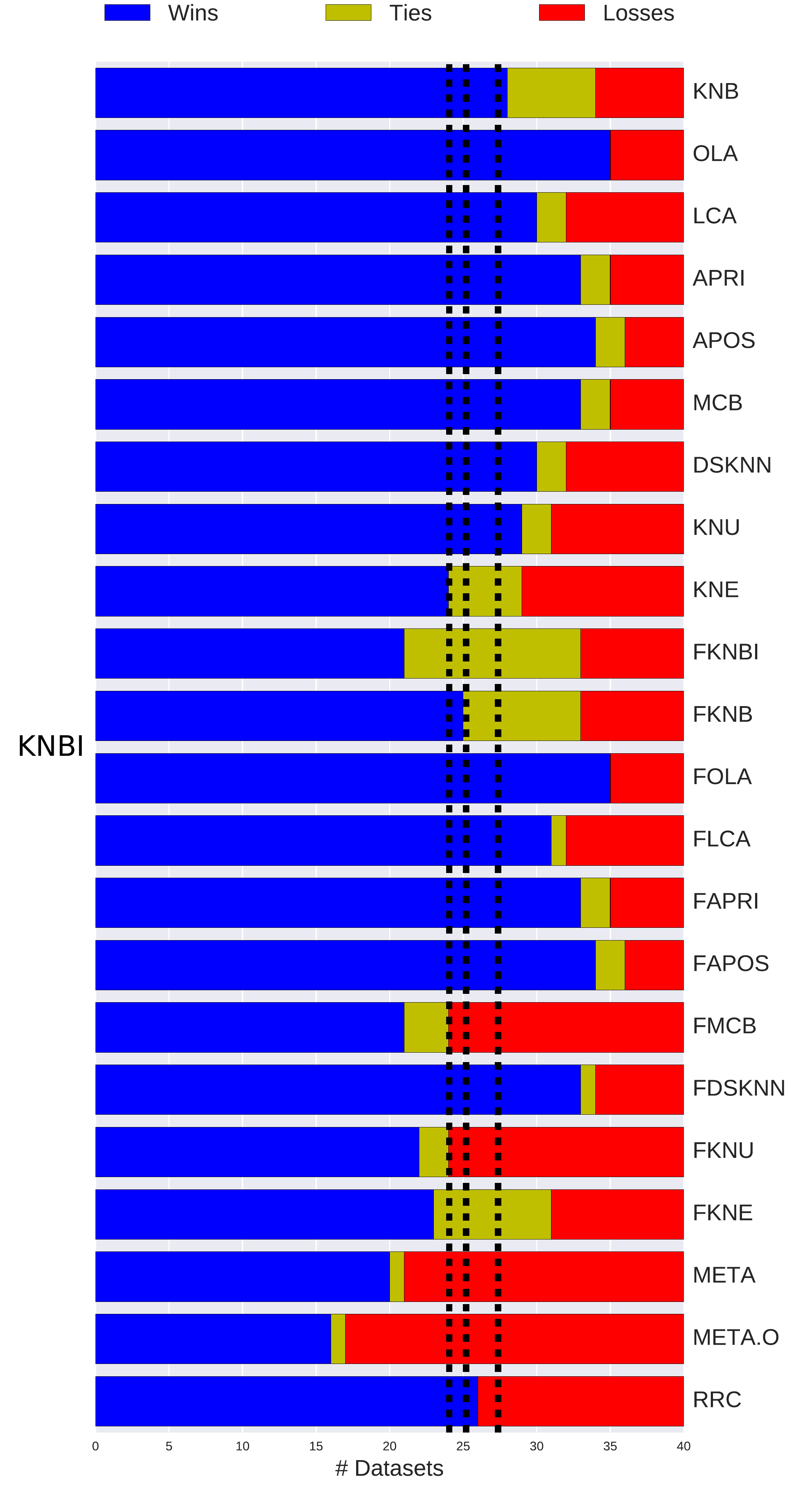}
        \label{fig:wtl_knbi_all}
    }
    \caption{
        Classification performance comparison of
        KNORA-B with different DES techniques
        (Figure \ref{fig:wtl_knb_all})
        and KNORA-BI with different DES techniques
        (Figure \ref{fig:wtl_knbi_all})
        in terms of wins, ties and losses considering the
        average AUC in the 40 datasets.
        The dashed lines (left to right) illustrates the
        critical values
        $n_c = \{24.05, 25.20, 27.37\}$ considering significance
        levels of $\alpha = \{0.10, 0.05, 0.01\}$, respectively.
    }
    \label{fig:wtlknbknbi}
\end{center}
\end{figure*}

Figure \ref{fig:wtlknbknbi} presents a pairwise comparison using the Sign Test \cite{signedtest:2006},
calculated using the wins, ties, and losses achieved
by the KNORA-B compared to other techniques (Figure \ref{fig:wtl_knb_all})
and
by the KNORA-BI compared to other techniques (Figure \ref{fig:wtl_knbi_all}).


For a comparison between one of the proposed techniques and a technique T,
the null hypothesis $H_0$ was that the proposed technique is not statistically different than T,
and a rejection of $H_0$ meant that the proposed technique is statistically better than T.
$H_0$ is rejected if the number of wins plus half of the number of ties
is greater or equal to $n_c$ (Equation \ref{eq:ncp_knb}):

\begin{equation}
    \label{eq:ncp_knb}
    n_c = \dfrac{n_{exp}}{2} + z_{\alpha} \times \dfrac{\sqrt[2]{n_{exp}}}{2}
\end{equation}

\noindent where $n_{exp} = 40$ (the number of experiments),
$n_c = \{24.05, 25.20, 27.37\}$, respectively for the levels of significance
$\alpha = \{0.10, 0.05, 0.01\}$.

Figure \ref{fig:wtl_knb_all} shows that, considering $\alpha = 0.05$,
KNORA-B statistically outperformed 7 out of the 8 regular DES techniques,
being only statistically equivalent to KNORA-E.
Comparing with FIRE-DES framework, KNORA-B outperformed only FAPOS (the worst of FIRE-DES in our experiments).
KNORA-B was not better than any of the state-of-art DES techniques.

On the other hand, Figure \ref{fig:wtl_knbi_all} shows that overall KNORA-BI statistically outperformed
a significant number of techniques studied (18 out of 22), considering a significance level $\alpha = 0.05$.
The only exceptions being the FMCB, FKNORA-U, META-DES, and META-DES.Oracle.  


\begin{figure}[h]
\centering
\includegraphics[scale=0.4]{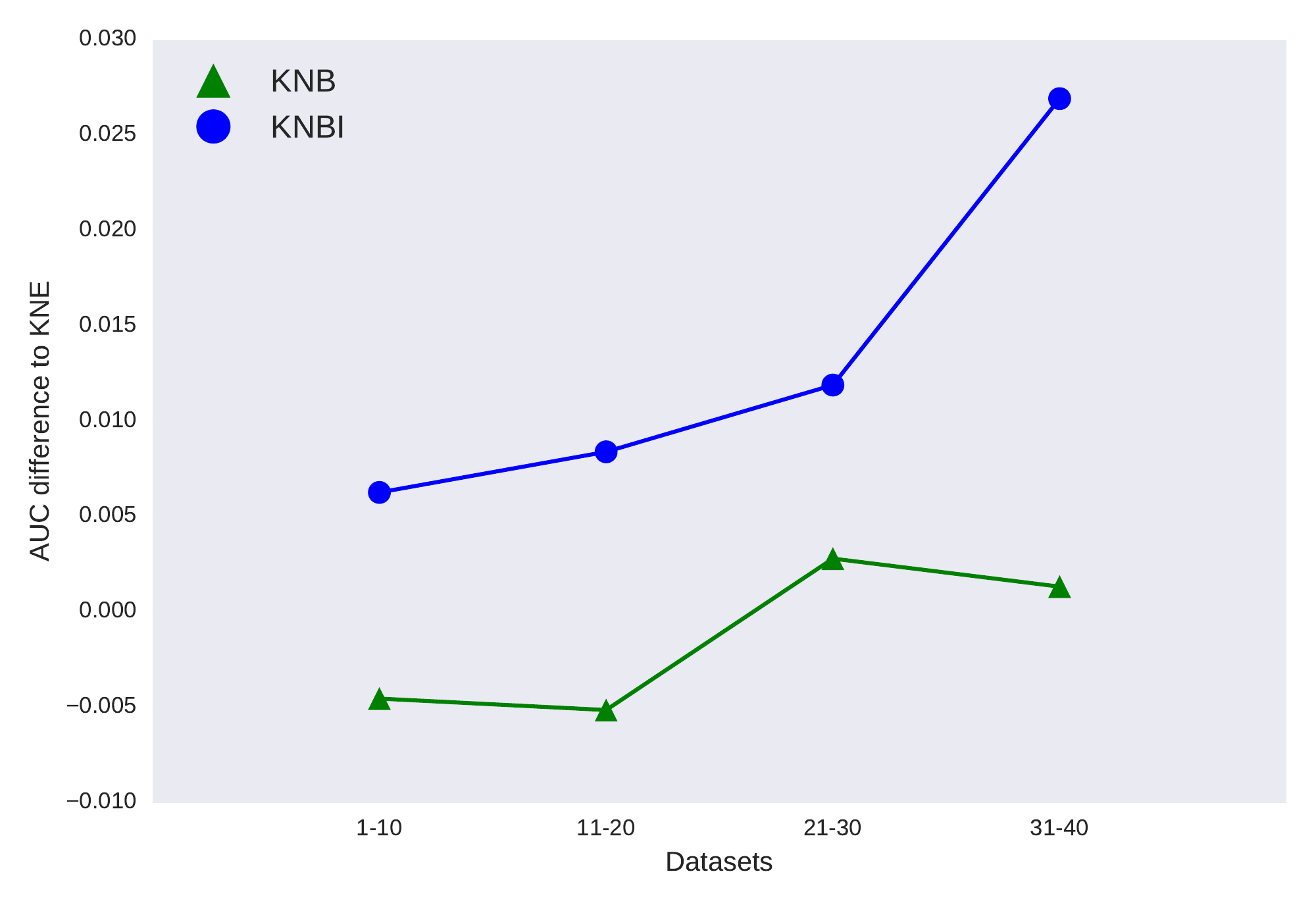}
\caption{AUC difference between KNORA-B and KNORA-E, and between KNORA-BI and KNORA-E.}
\label{fig:knbir}
\end{figure}

Figure \ref{fig:knbir} presents the average AUC difference from KNORA-B and KNORA-BI to
KNORA-E considering the datasets 1-10, 11-20, 21-30, and 31-40.
This figure shows that KNORA-B is better than KNORA-E optimally in 21-30
and slightly better in 31-40, while KNORA-BI was always better than KNORA-E
(the more the imbalance ratio the higher the difference).
This confirms that for all imbalance levels KNORA-BI was better than KNORA-E,
while KNORA-B was only better than KNORA-B for high imbalance level.

Based on the results, we can state with confidence
that KNORA-B achieved statistically equivalent performance of KNORA-E,
and KNORA-BI statistically outperformed KNORA-E and all other techniques
considered in this experiment.

\section{Conclusion}
\label{sec:conclusion}

In this paper, we proposed two DES techniques: K-Nearest Oracles Borderline
(KNORA-B) and K-Nearest Oracles Borderline Imbalanced (KNORA-BI).
KNORA-B is a DES technique based on KNORA-E that selects all
classifiers that correctly classify all samples in the region
of competence. If no classifier is selected, KNORA-B
reduces the region of competence maintaining the at least
one sample from each class in the original region of competence.
KNORA-BI is a variation of KNORA-B that reduces the region of
competence, but maintaining
at least one sample of the minority class (if such sample exists
in the original region of competence).

We conducted experiments using 40 datasets from the KEEL
software \cite{keel} with different levels of imbalance,
and compared KNORA-B and KNORA-BI with
8 regular DES techniques,
8 FIRE-DES techniques,
and 3 state-of-the-art DES techniques.

Results showed that KNORA-B had statistically equivalent performance of KNORA-E,
and KNORA-BI statistically outperformed KNORA-E in classification performance.
In fact, KNORA-BI achieved the best average classification performance over all
DES techniques considered in our experiments, including the state-of-the-art DES
techniques.

\section*{Acknowledgment}

The authors would like to thank
CAPES (Coordena\c{c}\~ao de Aperfei\c{c}oamento de Pessoal de N\'ivel Superior, in portuguese),
CNPq (Conselho Nacional de Desenvolvimento Cient\'ifico e Tecnol\'ogico, in portuguese)  and
FACEPE (Funda\c{c}\~ao de Amparo \`a Ci\^encia e Tecnologia do Estado de Pernambuco, in portuguese).



%

\bibliographystyle{IEEEtran}
\bibliography{IEEEabrv,content/bibliography/e2f-bibliography,content/bibliography/compbib}

\end{document}